\documentclass[conference]{IEEEtran}

\ifCLASSINFOpdf
\else
\fi
\usepackage{amsmath}
\usepackage{amsfonts}
\usepackage{algorithm}
\usepackage[noend]{algpseudocode}
\usepackage{amsmath}

\ifCLASSOPTIONcompsoc
  \usepackage[caption=false,font=normalsize,labelfont=sf,textfont=sf]{subfig}
\else
  \usepackage[caption=false,font=footnotesize]{subfig}
\fi
\usepackage{url}

\usepackage{color}
\usepackage{bbm}
\usepackage{amsfonts}
\usepackage{booktabs}   

\usepackage{gensymb}
\usepackage{pgfplots}
\pgfplotsset{compat=1.18}
\usepackage{tikz}
\usetikzlibrary{external}
\usepackage{graphicx} 

\makeatletter 
\newcommand{\linebreakand}{%
  \end{@IEEEauthorhalign}
  \hfill\mbox{}\par
  \mbox{}\hfill\begin{@IEEEauthorhalign}
}
\makeatother

\begin{document}
%
\title{A Reinforcement Learning Approach for Optimal Control in Microgrids}

\author{
\IEEEauthorblockN{1\textsuperscript{st} Davide Salaorni \\
\IEEEauthorblockA{
\textit{Politecnico di Milano}\\
Piazza L. Da Vinci, 32, Milan, Italy \\
\small\texttt{davide.salaorni@polimi.it}
}}
\and
\IEEEauthorblockN{2\textsuperscript{nd} Federico Bianchi\\
\IEEEauthorblockA{
\textit{Ricerca sul Sistema Energetico}\\
Via Rubattino, 54, Milan, Italy \\
\small\texttt{federico.bianchi@rse-web.it}
}}
\and
\IEEEauthorblockN{3\textsuperscript{rd} Marcello Restelli \\
\IEEEauthorblockA{
\textit{Politecnico di Milano}\\
Piazza L. Da Vinci, 32, Milan, Italy \\
\small\texttt{marcello.restelli@polimi.it} 
}}
\linebreakand
\IEEEauthorblockN{4\textsuperscript{st} Francesco Trovò \\
\IEEEauthorblockA{
\textit{Politecnico di Milano}\\
Piazza L. Da Vinci, 32, Milan, Italy \\
\small\texttt{francesco1.trovo@polimi.it} 
}}
}

\maketitle
\begin{center}
    {\small \textit{This paper has been accepted for presentation at IEEE IJCNN 2025. © 2025 IEEE. The final version will be published by IEEE.}}
\end{center}

\begin{abstract}
The increasing integration of renewable energy sources (RESs) is transforming traditional power grid networks, which require new approaches for managing decentralized energy production and consumption. Microgrids (MGs) provide a promising solution by enabling localized control over energy generation, storage, and distribution. This paper presents a novel reinforcement learning (RL)-based methodology for optimizing microgrid energy management. Specifically, we propose an RL agent that learns optimal energy trading and storage policies by leveraging historical data on energy production, consumption, and market prices. A digital twin (DT) is used to simulate the energy storage system dynamics, incorporating degradation factors to ensure a realistic emulation of the analysed setting. Our approach is validated through an experimental campaign using real-world data from a power grid located in the Italian territory. The results indicate that the proposed RL-based strategy outperforms rule-based methods and existing RL benchmarks, offering a robust solution for intelligent microgrid management.
\end{abstract}
\IEEEpeerreviewmaketitle

\vspace{-0.3cm}
\section{Introduction}

The process of producing, distributing, and consuming energy in Europe and in the US has undergone a significant paradigm change in the past years due to the rapid diffusion of renewable energy sources (RESs). The classical approach included a few large-scale producers supplying power to private individuals and companies, distributors that could effectively connect entities across the grid, and consumers whose role was limited to stipulating contracts for the energy supply. However, the decentralized nature of RESs challenges the traditional role of centralized power plants in maintaining grid reliability. In response, a policy shift has emerged, prioritizing the integration of distributed energy resources (DERs) within the so-called \emph{smart grid} structure, i.e., networks in which each node is allowed to produce, consume, store, and sell energy.

This shift reflects a recognition of DERs' potential to enhance grid resilience, promote energy independence, and seamlessly incorporate RESs into the system. 
However, to realize these benefits, it is crucial to develop optimized configurations and architectures for DERs, such as microgrids (MGs), which enable effective management and control of decentralized energy systems. An MG is a local power grid that can operate synchronously with the main electrical network and switch to a stand-alone (\emph{islanded}) mode without repercussions. Commonly, they are composed of different energy sources, e.g., microturbines, fuel cells, and photovoltaic panels, supplied with battery energy storage systems (BESSs), and designed to handle flexible loads~\cite{mariam2016}.

Even if MGs introduce a new level of flexibility, they also require tools for managing new elements to exploit the opportunities this setting provides. The need for intelligent management and control systems represents a significant challenge. Indeed, given the intrinsic combinatorial nature of the problem, designing a centralized controller located at the grid level to handle the production and consumption needs coming from each node, i.e., each MG, is no longer a viable option. Conversely, an alternative solution entails a distributed approach, where each MG defines its energy storage strategy.


According to the literature, a widely adopted MG control scheme is the hierarchical, multi-layered framework, which effectively manages operations by dividing tasks across three levels. Primary control ensures local stability and real-time power sharing; secondary control addresses voltage and frequency deviations while refining system performance; tertiary control focuses on optimizing the economic efficiency of the BESS and coordinating power exchange with the main grid~\cite{altin2021review}.
While the first two levels can be handled employing classical automation control strategies, for the third level, data-driven methods derived from the field of machine learning (ML), in particular, relying on deep reinforcement learning (DRL) techniques, have been proposed~\cite{dinata2024designing}. These methods have demonstrated the potential for learning optimal decision-making policies for MG control under conditions of uncertainty. Nonetheless, common available solutions optimize energy management based solely on costs and profits from MG interactions with the energy market, while neglecting the BESS degradation costs. Indeed, BESS replacement represents a substantial portion of MG's operating expenses. Therefore, literature works that disregard such a cost might provide significantly suboptimal policies in a real-world scenario.


Following this direction, our work aims to design a controller that optimizes the MG tertiary control level, i.e., to manage the BESS usage, avoid its excessive degradation, and reduce the expenses due to energy exchanges with the main grid.
In the current work, we propose:
\begin{itemize}
    \item an RL-based~\cite{Sutton1998} methodology to learn the optimal tertiary control strategy that uses historical information about energy consumption, power generation, market prices, and a digital twin (DT) to simulate the BESS;
    \item an experimental campaign to validate the proposed approach, which uses real-world data from the Italian energy market, real appliances consumption, photovoltaic production, and temperature profiles, to build effective data-driven control policies through RL approaches.
\end{itemize}

\section{Related Works} \label{sec:2}
In the past few years, a number of methods for controlling the tertiary level of MGs have been developed, especially in the field of energy storage management, spanning from RL techniques~\cite{dinata2024designing} to control theory methods~\cite{hu2021mpc}. 

Among the most relevant RL approaches, \cite{shojaeighadikolaei_weather-aware_2021} employed the Deep Q-Network (DQN) algorithm~\cite{mnih2013playing} to manage the load of an MG under energy generation uncertainties. The authors used RL to learn the strategy of a distributor agent, which has to set the energy price of the main grid and the policies of each independent \emph{prosumer} belonging to the MG. 
Authors of~\cite{kolodziejczyk_real-time_2021} employ DQN with real-world data of two simulation periods, i.e., a winter week and a summer week, corroborated by an exhaustive experimental campaign and hyperparameters tuning.
DQN is also adopted in~\cite{dominguez-barbero_optimising_2020}, where the objective is the optimization of isolated MG management without considering market dynamics, and in~\cite{chen_realistic_2019}, where authors focus on finding the best strategy with a realistic peer-to-peer energy trading model. 
Other researchers adopted policy-based methods~\cite{Peters2010PolicyGM}, which allow for continuous actions and the development of safer, more robust strategies. In~\cite{liu_deep_2021}, authors employ the Deep Deterministic Policy Gradient (DDPG) algorithm to tackle the mixed-integer linear programming of energy dispatch. Such a work considers multiple generation sources with the dual objective of satisfying power flow constraints and minimizing operational costs. 
In~\cite{guo_real-time_2022}, authors employ the Proximal Policy Optimization (PPO) algorithm to solve the optimal management of an MG comprising real-world data of forecasted generation sources, load consumptions, and energy prices. 

Regardless of the adopted RL methods, a common limitation in these approaches is related to the simplistic modeling of the BESS, which is often restricted solely to their electrical behavior. However, the BESS plays a main role within the MG due to its significant replacement cost. Therefore, not considering an accurate and realistic model for its calendar and usage degradation in the tertiary control strategies is a crucial weakness point of the above-mentioned methods.


A partial solution to these shortcomings is offered by~\cite{sui2020multi-agent}, which addresses the problem of scheduling charge, discharge, and resting periods with multiple batteries through a multi-agent approach. Since the study aims to minimize the system degradation, authors adopt an aging model influenced solely by temperature while ignoring other degradation factors related to usage, such as depth of discharge, state of charge (SoC), and current rates.
In~\cite{lin_-line_2021}, the authors solve the minimization of the MG operating costs under uncertain generation sources. Operational costs are considered to be related to MG power supply, battery usage, fuel price, and exchanges with the main grid. However, they do not provide information on how to set the parameters for the degradation cost in a real-world setting,
Nonetheless, authors of~\cite{mussi_reinforcement_2024} adopt a Thevenin equivalent circuit model to emulate the battery, alongside a circuit-based thermal model and a more sophisticated aging model to compute the system's degradation. However, they assume constant energy market prices and static ambient temperature, neglecting that they significantly influence MG performance, and formalize the problem with a discrete action space. Our work aims to build upon such an approach, addressing its limitations and tackling a more realistic scenario.

Finally, other works have been proposed based on the RL approach in the context of MG. Among the most relevant, the study in~\cite{syktiotis2024community} focuses on the optimization of electric vehicle charging through a multi-agent RL framework, while~\cite{hao2024lyapunov} proposes a safe RL method for the energy management problem to assure that training policy updates are confined within safety boundaries. Other studies handle the optimal power allocation problem, using either a primal-dual formulation~\cite{li2018distributed}, deep learning (DL)~\cite{guo2022training}, and RL~\cite{mu2024multi-objective} approaches.

\section{Problem Formulation} \label{sec:3}
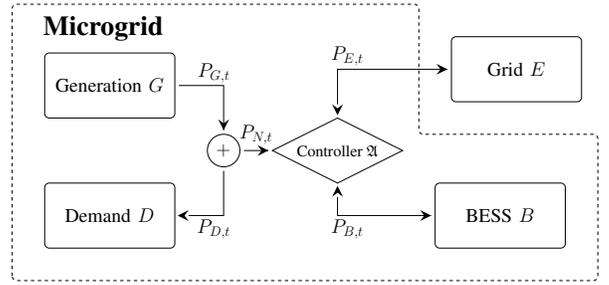
\begin{figure}[t]
    \centering
    \resizebox{\linewidth}{!}{\usetikzlibrary{arrows,decorations.markings}

\tikzset{myptrLeft/.style={decoration={markings,mark=at position 0 with {\arrow[scale=3,>=stealth]{<}}}, postaction={decorate}}}
\tikzset{myptrRight/.style={decoration={markings,mark=at position 1 with {\arrow[scale=3,>=stealth]{>}}}, postaction={decorate}}}
\tikzset{myptrBoth/.style={decoration={markings,mark=at position 0 with {\arrow[scale=3,>=stealth]{<}}, mark=at position 1 with {\arrow[scale=3,>=stealth]{>}}}, postaction={decorate}}}
    
\begin{tikzpicture}
    \centering

    \draw[rounded corners, dashed] (0,0) -- (0,8.5) -- (12.5,8.5) -- (12.5,4.5) -- (18,4.5) -- (18,0) -- cycle;
    \node[] at (2.8,7.75) {\Huge \textbf{Microgrid}} ;

    \draw[rounded corners] (1,5) rectangle (5,7);
    \node[align=center] at (3,6) {\LARGE Generation $G$} ;
    \node (generation) at (5,6) {}; 

    \draw[rounded corners] (1,1) rectangle (5,3);
    \node[align=center] at (3,2) {\LARGE Demand $D$} ;
    \node (demand) at (5.3,2) {}; 

    \draw[] (6.5,4) circle (0.5);
    \node[align=center] at (6.5,4) {\LARGE $+$} ;
    \node (sumUp) at (6.5,4.5) {}; 
    \node (sumDown) at (6.5,3.5) {}; 
    \node (sumRight) at (7,4) {}; 

    \draw [] (8, 4) -- (10,5) -- (12,4) -- (10,3) -- cycle;
    \node[align=center] at (10,4) {\Large Controller $\mathfrak{A}$} ;
    \node (controllerUp) at (10,5.3) {};
    \node (controllerDown) at (10,2.7) {};
    \node (controllerLeft) at (8,4) {};

    \draw[rounded corners] (13,1) rectangle (17,3);
    \node[align=center] at (15,2) {\LARGE BESS $B$} ;
    \node (battery) at (13,2) {};

    \draw[rounded corners] (13.5,5.5) rectangle (17.5,7.5);
    \node[align=center] at (15.5,6.5) {\LARGE Grid $E$} ;
    \node (grid) at (13.5,6.5) {};

    \draw [myptrRight] (generation) -| (sumUp);
    \draw [myptrLeft] (demand) -| (sumDown);
    \draw [myptrRight] (sumRight) -- (controllerLeft);
    \draw [myptrBoth] (controllerUp) -- (10,6.5) -- (grid);
    \draw [myptrBoth] (controllerDown) -- (10,2) -- (battery);

    \node[] at (6.2, 6.4) {\LARGE $P_{G,t}$};
    \node[] at (6.2, 1.6) {\LARGE $P_{D,t}$};
    \node[] at (7.5, 4.5) {\LARGE $P_{N,t}$};
    \node[] at (10.3, 6.9) {\LARGE $P_{E,t}$};
    \node[] at (10.3, 1.6) {\LARGE $P_{B,t}$};
    
\end{tikzpicture}
}
    \caption{Schema of the interactions between core MG elements. Single-edge arrows indicate a unidirectional positive exchange, while double-edge arrows represent bidirectional ones.}
    \label{fig:control_schema}
    \vspace{-0.5cm}
\end{figure}

Our work aims to design a controller $\mathfrak{U}$ to optimize the power flow in an MG setting while minimizing costs related to the degradation of the BESS and energy exchanges with the main grid. As depicted in Figure~\ref{fig:control_schema}, an MG is a node connected to a broader grid $E$ and comprises an overall generation source $G$, an overall consumption $D$, and a BESS $B$.
In such an architecture, the controller serves as the central element that coordinates the entire system and regulates how to manage the electrical power flowing through the MG.

In this setting, at a specific time $t$, power production primarily relies on renewable generation sources, which generate an inherently uncertain amount of power, denoted as $P_{G,t}$. Meanwhile, the energy required to power the MG's electrical devices is aggregated within the total demand $D$, with $P_{D,t}$ representing the power consumed by MG users at time $t$. Ideally, power generation would fully meet consumption demands. However, in practice, these two signals often diverge both in terms of magnitude and in their temporal patterns. 
Indeed, power generated by RESs, e.g., photovoltaic panels, strictly depends on weather conditions, season, and time of the day. On the other hand, consumption profiles follow users' behavior within a residential context~\cite{kong2019short-term}, and they do not synchronize well with generation patterns. Thus, the difference between production and consumption generates the net power:
\begin{equation}
    P_{N,t} = P_{G,t} - P_{D,t},
\end{equation}
either positive or negative, which must be managed effectively by the controller.
More specifically, if $P_{N,t}$ is a negative quantity, meaning that the produced energy cannot satisfy the demand, the controller has to decide whether to retrieve the remaining amount from the battery or the grid. To do so, it selects $P_{B,t}$, i.e., the fraction of $P_{N,t}$ that it wants to retrieve from the battery, while the remaining part $P_{E,t}$ is acquired from the main grid. Conversely, if $P_{N,t}$ is positive, it means that there is a surplus of power with respect to the users' demand. In this case, the controller action is intended as the proportion of net power $P_{N,t}$ to store within the battery $P_{B,t}$, while the remaining part $P_{E,t}$ will be sold to the grid. Formally, at each time $t$, the controller has to choose the value $a_t$ with $a_t \in [0,1]$ such that:
\begin{align}
    P_{B,t} & = a_t P_{N,t} \,\\
    P_{E,t} & = (1 - a_t) P_{N,t} \,.
\end{align}
The chosen action should be compliant with the physical constraints of the BESS since it is not possible to overcharge or excessively drain it, and there exist physical limitations on the amount of power it can exchange over a limited time period. 
Let us define the storage unit state of charge (SoC) at time $t$:
\begin{equation}
    \sigma_t := \sigma_{1} + \sum_{h=2}^t \frac{i_h \Delta \tau}{C_{h}},
\end{equation}
where $\sigma_{1} \in \mathbb{R^+}$ is the value for the SoC at initial time, $C_{h}$ and $i_h$ are the maximum internal capacity in [\textit{Ah}] and the current in [\textit{A}] of the energy storage unit at time $h$, and $\Delta \tau$ is the considered interval of time in [\textit{h}]. To ensure compliance with these constraints, the power $P_{B,t}$ must be bounded by: 
\begin{align}
        P_{dch} &\leq P_{B,t} \leq P_{ch}, \label{eq:constr1}\\ 
        \frac{(\sigma_{min} - \sigma_{t})}{\Delta \tau} C_{t} V_t &\leq P_{B,t} \leq \frac{(\sigma_{max} - \sigma_t)}{\Delta \tau}C_{t} V_t \label{eq:constr2}
\end{align}
where $P_{dch} < 0$ and $P_{ch} > 0$ denote the maximum discharging and charging power, respectively, $\sigma_{min} \in \mathbb{R^+}$ and $\sigma_{max} \in \mathbb{R^+}$ are the minimum and maximum values for the SoC, and $V_t$ is the voltage in [\textit{V}] at time $t$, respectively.

\subsection{Controller's goal}

The goal of the controller is to learn the optimal strategy leading to the maximization of economic profits derived from the management of the MG. Specifically, two main components arise in the optimization of the profits/losses: the gains/costs due to trading with the energy market and the costs related to BESS degradation. Formally, the controller wants to maximize over a time horizon $\mathcal{T}$ the total profit obtained following a specific strategy, a.k.a.~policy, $\pi = (a_0, \ldots, a_{\mathcal{T}})$. Hence, the \emph{total profit} of the MG is defined as:
\begin{equation}
    R_{\mathcal{T}}(\pi) = \sum_{t=1}^\mathcal{T} \ [r_{trad}(a_t) + r_{deg}(a_t)],
\end{equation}
where $r_{trad}(a_t) \in \mathbb{R}$ is the reward/cost gained from the exchanges of energy the MG makes with the market, and $r_{deg}(a_t) < 0$ is the cost due to battery degradation.

Formally, the trading component $r_{trad}(a_t)$ of the profit is:
\begin{align}
    r_{trad}(a_t) = (p_t^{sell} P^+_{E,t} + p_t^{buy} P^-_{E,t}) \Delta \tau \,, \label{eq: r_trad}
\end{align}
where $p^{sell}_{t}$ and $p^{buy}_t$ are the unit energy prices for selling and buying energy at time $t$, respectively ($p^{sell}_{t} < p^{buy}_t$), and $P^+_{E,t}$ and $P^-_{E,t}$ are the positive and negative part of $P_{E,t}$, respectively.\footnote{Given a quantity $q \in \mathbb{R}$, we define its positive part as $q^+ := \max\{0, q \}$ and its negative part as $q^- := \min \{0, q \}$.}

The degradation term $r_{deg}(a_t)$ depends on the BESS's State of Health (SoH), which is related to its aging. Indeed, an energy storage system starts its life cycle at the maximum SoH, i.e., $100$\%, which monotonically decreases until the end-of-life (EOL) value, indicating that the system is no longer capable of performing the current task (usually between $80-60$\% depending on the specific application). Formally, the SoH, indicated as $\rho_t \in [0,1]$, at time $t$ is defined as:
\begin{equation}
    \rho_t := \frac{C_{t}}{C_N},
\end{equation}
where $C_N \in \mathbb{R}^+$ is the nominal internal capacity of the system in [\textit{Ah}]. The degradation of the BESS, in terms of $C_t$, will depend both on its usage and calendar aging. Thus, the degradation cost term can be expressed as:
\begin{equation}
    r_{deg}(a_t) = \frac{\rho_t - \rho_{t-1}}{1-\rho_{EOL}} \mathcal{R} \,, \label{eq: r_deg}
\end{equation}
where $\rho_{t}$ and $\rho_{t-1}$ are the battery SoH values at time $t$ and $t-1$, respectively ($\rho_{t} < \rho_{t-1}$), $\rho_{EOL} \in (0, 1)$ is the end of life (EOL) SoH, and $\mathcal{R}$ is the replacement cost for the BESS.

\section{Proposed Method} \label{sec:4}
In this section, we formalize the presented problem as a \emph{Markov Decision Process} (MDP), a mathematical framework useful to solve sequential decision-making problems, and propose an RL-based framework using the MDP and a digital twin for the BESS component of the MG to find the optimal control strategy.

More specifically, an MDP models the process of sequential decision-making over a dynamic environment and is formally defined as a tuple:
\begin{equation}
    \mathcal{M} = (S, A, P, R, \gamma, \mu_0),
\end{equation}
where $S$ is the set of the vectors encoding the state of the environment, $A$ is the set of actions the controller can perform in each state $s \in S$, $P$ is the transition matrix, specifying the probability of moving to a state $s'$ starting from state $s$ and performing action $a \in A$ with $s,s' \in S$, $R \in \mathbb{R}^+$ is the instantaneous reward provided by performing action $a$ in state $s$, $\gamma \in [0, 1]$ is a discount factor, that allows designing the problem fostering myopic or farsighted strategies, and $\mu_0$ is a distribution over the states $S$ stating where the decision-making process starts.

In our setting, $P$ and $R$ are unknown due to the stochasticity of the market prices and the energy generation sources. The goal of an RL procedure is to learn a policy $\pi$ to maximize the cumulative long-term reward $R_{\mathcal{T}}(\pi)$, using the data generated by the MDP. In particular, at a specific time $t+1$, the RL learning algorithm will process the following structure of data:
\begin{equation}
    (s_t, a_t, r_t, s_{t+1}),
\end{equation}
where $s_t \in S$ is the current state, $a_t \in A$ is the action performed in $s_t$, $r_t$ is the reward gathered from the environment at state $s_t$ by doing action $a_t$, and $s_{t+1}$ is the next state. In the following section, we will show our design for such elements to solve the problem of controlling an MG.

\subsection{State space}
The state space vector $s_t$ comprises the signals that the agent receives from the environment at each time step $t$. Formally:
\begin{align}
    s_t = &\left( \sigma_t, T_t, \widehat{P}_{D,t}, \widehat{P}_{G,t}, p_t^{buy}, p^{sell}_t, \right.\nonumber\\
    & \,\,\, \left. \text{cos}(\varphi^d_t), \text{sin}(\varphi^d_t),
    \text{cos}(\varphi^y_t), \text{sin}(\varphi^y_t) \right),
\end{align}
where:
\begin{itemize}
    \item $\sigma_t$ is the storage unit SoC at time $t$;
    \item $T_t$ is the current battery temperature;
    \item $\widehat{P}_{D,t}$ is the estimate of energy demand $P_{D,t}$ at time $t$;
    \item $\widehat{P}_{G,t}$ is the estimate of energy generation $P_{G,t}$ at time $t$;
    \item $p^{buy}_t$ and $p^{sell}_t$ are the buying price and the selling price of the energy from the market at time $t$, respectively;
    \item $\varphi^d_t \in [0, 2\pi] $ is a variable representing the angular position of the clock in a day, given by $\varphi^d = \frac{2\pi \tau_d}{\mathcal{T}_d}$, where $\tau_d \in [0, \mathcal{T}_d]$ is the current time of the day in seconds and $\mathcal{T}_d$ is the total number of seconds in a day;
    \item $\varphi^y_t \in [0, 2\pi]$ is a variable representing the angular position of the current time over the entire year, given by $\varphi^y = \frac{2\pi \tau_y}{\mathcal{T}_y}$, where $\tau_y \in [0, \mathcal{T}_y]$ is the current time in seconds and $\mathcal{T}_y$ is the total number of seconds in a year.
\end{itemize}

Some remarks are in order. First, the battery SoC $\sigma_t$ and temperature $T_t$ variables represent the current state of the BESS component. Their joint evolution determines the degradation of the system~\cite{xu2018modeling}. 
Second, $\widehat{P}_{D,t}$, $\widehat{P}_{G,t}$, $p^{buy}_t$, and $p^{sell}_t$ enclose the information related to the exogenous factors influencing the system, i.e., energy consumption, generation, and buying/selling price. Ideally, the decision should be driven by the demand $P_{D,t}$ and generation $P_{G,t}$ at time $t$, which are unknown at the time the agent is making a decision. In our approach, we use the corresponding values at time $t-1$ since they represent a sufficiently accurate estimate as $\widehat{P}_{D,t}$ and $\widehat{P}_{G,t}$.\footnote{More complex approaches can be used to get better approximation of such profiles, e.g., using the techniques presented in~\cite{rastkar2024improving}}. Conversely, we assume that the energy prices $p^{buy}_t$ and $p^{sell}_t$ are known in advance since, commonly, the energy market deals are decided one day ahead.
Finally, the observation space is enriched with a specific transformation of the current time variables $\varphi^d$ and $\varphi^y$ to take into account the day-night and seasonal periodicity related to the intra-day and seasonal variability of both energy sources and MG demands. 

\subsection{Action space}

The action space in our setting is the continuous action variable $a_t \in [0, 1]$, representing the proportion of energy to \textit{dispatch} (\textit{take}) to (from) the BESS.
Depending on the sign of $P_{N,t}$, the action $a_t$ has a different meaning. If $P_{N,t} > 0$, and therefore $P_{B,t} \geq 0$ and $P_{E,t} \geq 0$, it regulates the proportion between the power used to charge the battery and the one sold to the main grid. Conversely, if $P_{N,t} < 0$, the action $a_t$ regulates the proportion of demanded energy that will be taken from the energy storage and the one bought from the market, thus $P_{B,t} \leq 0$ and $P_{E,t} \leq 0$.

We remark that $P_{B,t}$ has to adhere to constraints described in Equations~\eqref{eq:constr1} and~\eqref{eq:constr2}. Hence, the chosen action must be clipped to feasible values so as not to actuate unfeasible or harmful actions. Such a restriction avoids actions that otherwise would be disallowed by the battery management system (BMS), the component in charge of system safety, and fosters the agent to learn within a feasible operating region.

\subsection{Reward}
We designed a reward such that the feedback provided by the environment can help the agent evaluate whether the action taken was beneficial. Formally, we considered the instantaneous reward:
\begin{equation}
    r_t = [r_{trad}(a_t) + r_{op}(a_t)] + \lambda r_{clip}(a_t) \label{eq:r_t},
\end{equation} 
where $r_{clip}(a_t)$ is a penalty term, detailed in the following, to discourage the learner from doing actions $a_t$ that are not compliant with the environment, and $\lambda$ is a weight to regulate how much such component should be influencing the learning procedure.
We recall that the first two elements are the same as we defined in Equations~\eqref{eq: r_trad} and \eqref{eq: r_deg}, i.e., $r_{trad}(a_t)$ is the revenue/cost due to the interaction with the energy market, and $r_{deg}(a_t)$ is the cost due to the utilization and degradation of the battery.
The term $r_{clip}(a_t)$ is a penalty given to the agent performing an action that needs to be clipped for structural constraints, i.e., that are violating the constraints in Equations~\eqref{eq:constr1} and~\eqref{eq:constr2}.\footnote{In the following formulation, $r_{clip}(a_t)$ is computed only w.r.t.~to constraint in Equation~\eqref{eq:constr2}, which in our setting is stricter than Equation~\eqref{eq:constr1}.} Formally, it is defined as: 
\begin{align}
    r_{clip,t} = - & \max \Big\{0, a_t P_{N,t} + \frac{(\sigma_t - \sigma_{\max})}{\Delta \tau} C_t V_t, \nonumber\\
    &\frac{(\sigma_{\min} - \sigma_t)}{\Delta \tau} C_t V_t - a_t P_{N,t} \Big\}\,.
\end{align}

\subsection{Learning procedure}
The learning procedure is based on collecting samples $(s_t, a_t, r_t, s_{t+1})$ from the interaction between a learning agent $\mathfrak{U}$ and the environment. The pseudocode of the exchanges occurring during the training procedure is provided in Algorithm~\ref{alg:psueducode}.

\begin{algorithm}[th]
\caption{Interaction between Agent and Environment}
\label{alg:psueducode}
\begin{algorithmic}[1]
\State \textbf{Initialize:} $s_0$, $\{ \mathcal{P}_D^{(i)}\}_{i=1}^M$, $\mathcal{P}_{G}$, $\mathcal{C}_{buy}$, $\mathcal{C}_{sell}$, $\mathcal{K}$, $B(\cdot)$, $\pi (\cdot)$

\For{$j \in \{1, \ldots, n_{ep} \}$ }
    \State Sample demand profile $\mathcal{P}^{(i)}_{D} \sim Unif\big(\{ \mathcal{P}_D^{(i)}\}_{i=1}^M\big)$\label{line:3}
    \State Initialize $\sigma_1$, $T_1$, $\rho_1$ \label{line:4}
    \For{$t \in \{1, \ldots, \mathcal{T} \}$}
        \State Compute estimates $\widehat{P}_{G,t}, \widehat{P}^{(i)}_{D,t}$ \label{line:6}
        \State Observe current state $s_t$ \label{line:7}
        \State Agent takes action $a_t \sim \pi(s_t)$ \label{line:8}
        \State Compute $P_{B,t} \gets a_t (P_{G,t} - P^{(i)}_{D,t})$ \label{line:9}
        \State Update $(\sigma_{t+1}, T_{t+1}, \rho_{t+1}) \leftarrow B(\sigma_t, T_t, K_t, P_{B,t})$ \label{line:10}
        \State Compute $P_{E,t} \gets (1 - a_t) (P_{G,t} - P^{(i)}_{D,t})$ \label{line:11}
        \State Collect reward $r_t$ \label{line:12}
        \State Update policy $\pi(\cdot)$ \label{line:13}
    \EndFor
\EndFor
\end{algorithmic}
\end{algorithm}

As input, the algorithm requires an initial state of the microgrid $s_0 \in S$, values for the time horizon $\{1, \ldots, \mathcal{T} \}$ of a set of demand profiles $\{ \mathcal{P}_D^{(i)}\}_{i=1}^M$, with $M \in \mathbb{N}$, a generation profile $\mathcal{P}_G := (P_{G,1}, \ldots P_{G,\mathcal{T}})$, market selling and buying prices sequences $\mathcal{C}_{sell} := (p^{sell}_1, \ldots, p^{sell}_ {\mathcal{T}})$ and $\mathcal{C}_{sell} := (p^{buy}_1, \ldots, p^{buy}_{\mathcal{T}})$ (respectively), and the ambient temperature profile $\mathcal{K} := (K_1, \ldots, K_{\mathcal{T}})$.
We remark that we use multiple demand profiles since each load is specific for an MG. More specifically, a demand profile is formally defined as $\mathcal{P}_D^{(i)} := (P_{D,1}^{(i)}, \ldots P_{D,\mathcal{T}}^{(i)})$.
Conversely, we used a single generation, market price, and ambient temperature sequence, assuming the simulated MGs have similar locations. We also need a model for the BESS $B(\sigma_t, T_t, K_t, P_{B,t})$ which, given a SoC $\sigma_t$, the internal temperature $T_t$, the ambient temperature $K_t$, and a charge/discharge power $P_{B,t}$, is able to provide the SoC $\sigma_{t+1}$, its internal temperature $T_{t+1}$ and the degradation occurred in the last step $\rho_{t+1}$.
Finally, we require an initial policy $\pi(\cdot)$ that will be optimized over the procedure.

The learning process occurs over $n_{ep}$ episodes, along which we simulate the MG by selecting each time a specific demand profile $\mathcal{P}^{(i)}_D$ uniformly at random, considered for the entire training episode (Line~\ref{line:3}).
Once the episode is set and the battery is initialized (Line~\ref{line:4}), we simulate a step $t$ of the interaction between the environment and the agent.\footnote{
In this section, we consider an arbitrary step $t$. In practice, the control of microgrid occurs at a frequency in the interval $(0.0003, 0.0011)$ Hz \cite{mussi_reinforcement_2024}.} At first, the agent estimates the value of the next generation $\widehat{P}_{G,t}$ and demand $\widehat{P}_{D,t}$ (Line~\ref{line:6}), to build the current state of the environment $s_t$ (Line~\ref{line:7}). Based on the state, the agent executes the action $a_t \sim \pi(s_t)$ prescribed by the policy (Line~\ref{line:8}). Such an action determines the energy to be sent to the BESS $P_{B,t}$ (Line~\ref{line:9}), and its corresponding evolution (Line~\ref{line:10}), and the energy to exchange with the main grid $P_{E,t}$ (Line~\ref{line:11}). Based on the degradation and the exchanges with the main grid, the agent receives the feedback $r_t$, as defined by Equation~\eqref{eq:r_t} (Line~\ref{line:12}). Finally, the agent updates its policy $\pi(\cdot)$ (Line~\ref{line:13}). 
Common choices for the algorithms able to perform such an update are detailed in~\cite{Peters2010PolicyGM}.
We remark that while the environment is reset at each episode, the policy $\pi(\cdot)$ is kept fixed between episodes so that the information incorporated in previous episodes is retained.


\section{Experimental evaluation} \label{sec:5}

In this section, we validate the presented approach in a real-world scenario based on an MG for household use in the Italian region. We leverage data coming from $2015$ to $2020$, using the first $4$ years for training the agent $\mathfrak{U}$ and the last one for testing its efficacy. We used an energy generation profile of a single photovoltaic residential implant of $3$kW from~\cite{pfenninger2016longterm} and a daily ambient temperature profile from~\cite{staffell2023global}, both providing averaged measurements across the country. Energy market prices are computed using the Italian Gestore Mercati Energetici (GME)~\cite{GME}, providing hourly data of the energy National Single Price and accounting for fixed tax-related and managing costs. 
Finally, we considered a total of $398$ consumption profiles as realistic samples of Italian household consumption across the North, Center, South, and islands of the country from~\cite{fioriti2022optimal}. We randomly split the profiles into training and test sets, with $M = 370$ and $N = 28$ profiles respectively. Experiments have been conducted on a MacBook Pro, $2023$, with an Apple M2 Pro CPU, $16$ GB of RAM, and MacOS Sequoia $15$. The codebase will be available after the review phase to avoid authors' information disclosure.

\subsection{Analysed Methods and Performance Metrics}

In our solution, from now on denoted as ${RL^{\star}}$, the RL agent is trained on $4$ parallel environments leveraging the PPO algorithm~\cite{raffin2021stable-baselines3}, a policy-based method that allows using a continuous action space and has shown its effectiveness in multiple applicative fields~\cite{depaola2024power,yu2022suprising}. The training phase lasts $n_{ep} = 100$ episodes and has a decision step of $\Delta \tau = 3600s$ (or equivalently a frequency of $0.0003$ Hz). For the networks of PPO's actor and critic has been chosen a feedforward architecture with $2$ hidden layers of $64$ neurons each. The codebase of the framework is available here: \url{https://github.com/Daveonwave/ErNESTO-gym}.\footnote{More details about the hyperparameters selected to run the experiments are reported in the repository.}

In our solution, we simulate the dynamics of the BESS using the ErNESTO-DT~\cite{salaorni2023ernesto}, which emulates the evolution of battery SoC, temperature, and SoH over time. It models a realistic battery pack built by connecting multiple lithium-ion NMC cells to achieve a nominal voltage of $350$ V and an energy content of $21$ kWh.
We assume a battery replacement cost of $\mathcal{R} =3000$€.

We compared ${RL^{\star}}$ with common deterministic strategies and state-of-the-art RL-based controllers of MG energy storage systems. More specifically, we evaluated \emph{X-Y}s: deterministic rule-based strategies that dispatch $X$\% of $P_{N,t}$ to the battery and the remaining $Y$\% to the grid. Specifically, we tested $20$-$80$, $50$-$50$, and $80$-$20$; \emph{OnlyGrid (OG)}: a rule-based strategy forcing the interaction with the main grid without using the battery. This corresponds to the $0$-$100$ deterministic policy; \emph{BatteryFirst (BF)}: a rule-based policy fostering battery usage as much as possible before interacting with the main grid. It corresponds to $100$-$0$ deterministic policy; \emph{RL-base}: baseline reproducing the solution conceived by~\cite{mussi_reinforcement_2024}, in which the training is done with fixed ambient temperature and market prices. Market prices are computed as an average over the prices of the training years, while ambient temperature has been fixed to $K_t = 25\degree C$; \emph{RL-base+}: an extension of \emph{RL-base} using fixed ambient temperature but incorporating information about the evolving market dynamics.
We compare the different policies $\mathfrak{U}$ in terms of average (over the validation profiles) empirical reward over the time horizon $t \in \{1, \ldots, \mathcal{T} \}$:
\begin{equation}
    \widehat{R}_t(\mathfrak{U}) = \frac{1}{N} \sum_{i=1}^N \sum_{h=1}^t [r_{trad}^{(i)}(a_h) + r_{deg}^{(i)}(a_h) ],
\end{equation}
where $r_{trad}^{(i)}(a_h)$ and $r_{deg}^{(i)}(a_h)$ are the trading and degradation costs corresponding to the $i$-th demand profile $\mathcal{P}^{(i)}_D$ at time $t$. We also compared the separate contribution of trading and degradation to the final reward as follows:
\begin{equation}
    \widehat{R}_{\square,t}(\mathfrak{U}) = \frac{1}{N} \sum_{i=1}^N \sum_{h=1}^t r_{\square}^{(i)}(a_h)
\end{equation}
with $\square \in \{trad, deg\}$. In particular, we evaluate the gap $\Delta \widehat{R}_{\square,t}(\mathfrak{U}, \mathcal{B})$ w.r.t.~a baseline method $\mathcal{B}$:
\begin{equation}
    \Delta \widehat{R}_{\square,t}(\mathfrak{U}, \mathcal{B}) := \widehat{R}_{\square,t}(\mathfrak{U}) - \widehat{R}_{\square,t}(\mathcal{B}).
\end{equation}
Positive values for $\Delta \widehat{R}_{\square,t}(\mathfrak{U},\mathcal{B})$ corresponds to larger performance of $\mathfrak{U}$ than $\mathcal{B}$.

\begin{figure}[th!]
    \vspace{-0.3cm}
    \centering
    \includegraphics[width=\linewidth]{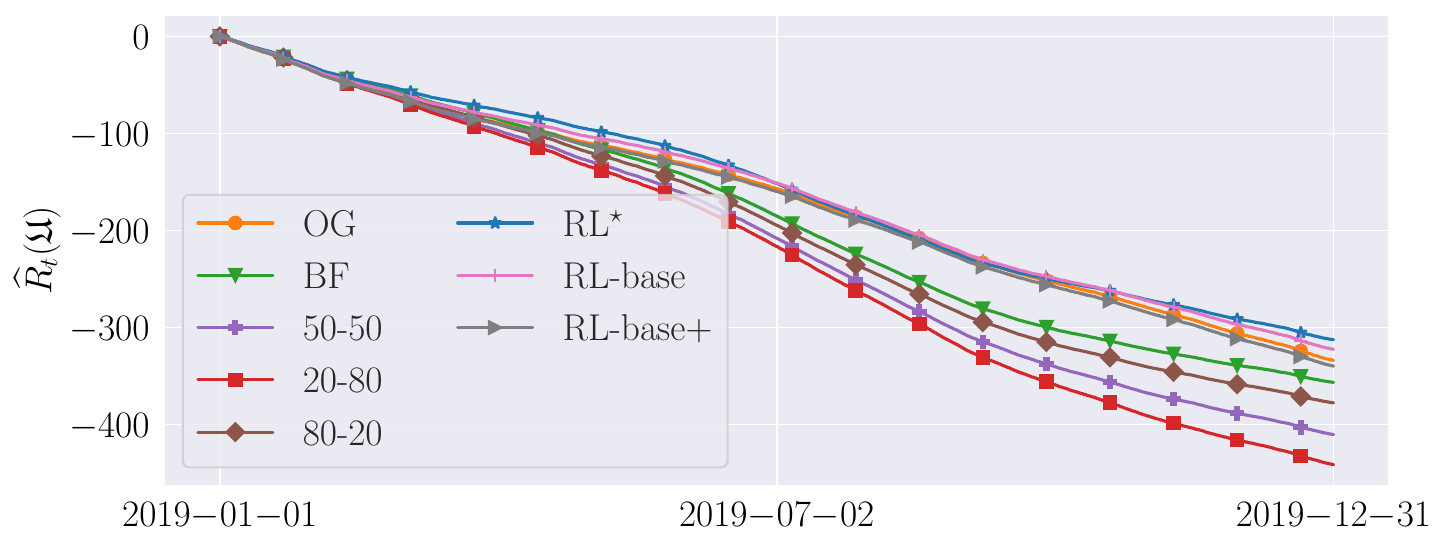}
    \vspace{-0.5cm}
    \caption{Cumulative average empirical reward.}
    \label{fig:average_return}
    \vspace{-0.5cm}
\end{figure}

\subsection{Campaign results}

Figure~\ref{fig:average_return} reports the cumulative empirical reward $\widehat{R}_t(\mathfrak{U})$ of the policies corresponding to the analysed control strategies over the time horizon.
It is worth noting that, on average, none of the strategies achieve a positive total return. This was expected since the cost of buying energy is always larger than the selling price, and the generation profile produces energy in excess of the demand only $39.6$\% of the times over the year of testing. Moreover, the reward also includes the degradation component, which is strictly negative since it is related to BESS usage.

Among the analysed approaches, the proposed $RL^\star$ solution provides the best average results over the time horizon. Within the central months of the year (approximately June to September), the performances have been on par or slightly worse than those of previous RL-based solutions and OG. However, at the end of the time horizon, the $RL^\star$ method reduces the cost of $3.2\%$ w.r.t.~\emph{RL-base}, i.e., the state-of-the-art RL solution, and of $6.7\%$ w.r.t.~OG, i.e., the best among the rule-based strategies. The strong OG performance, in this scenario, is primarily due to the relatively high cost of BESS replacement, which in turn brings high degradation costs for those methods that rely more heavily on BESS utilization. 

\begin{figure}[th!]
    \vspace{-0.4cm}
    \centering
    \begin{minipage}{\linewidth}
        \centering
        \subfloat[]{
            \includegraphics[width=\linewidth]{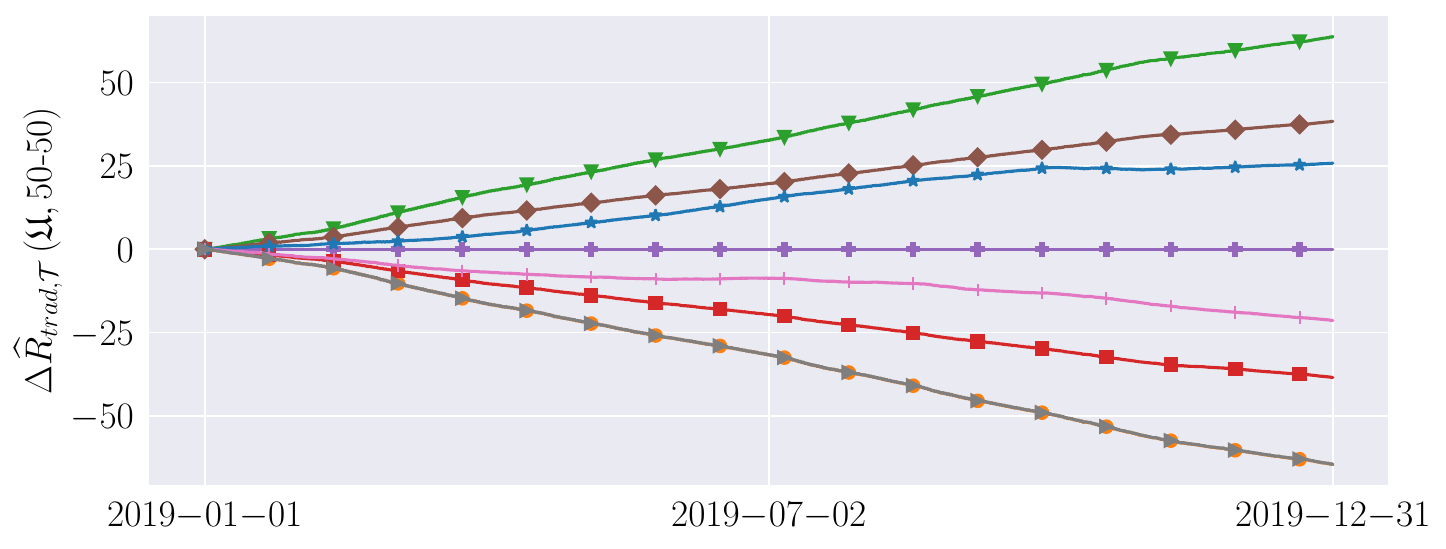}
            \label{fig:r_trad}}
    \end{minipage}
    \begin{minipage}{\linewidth}
        \centering
        \subfloat[]{
            \includegraphics[width=\linewidth]{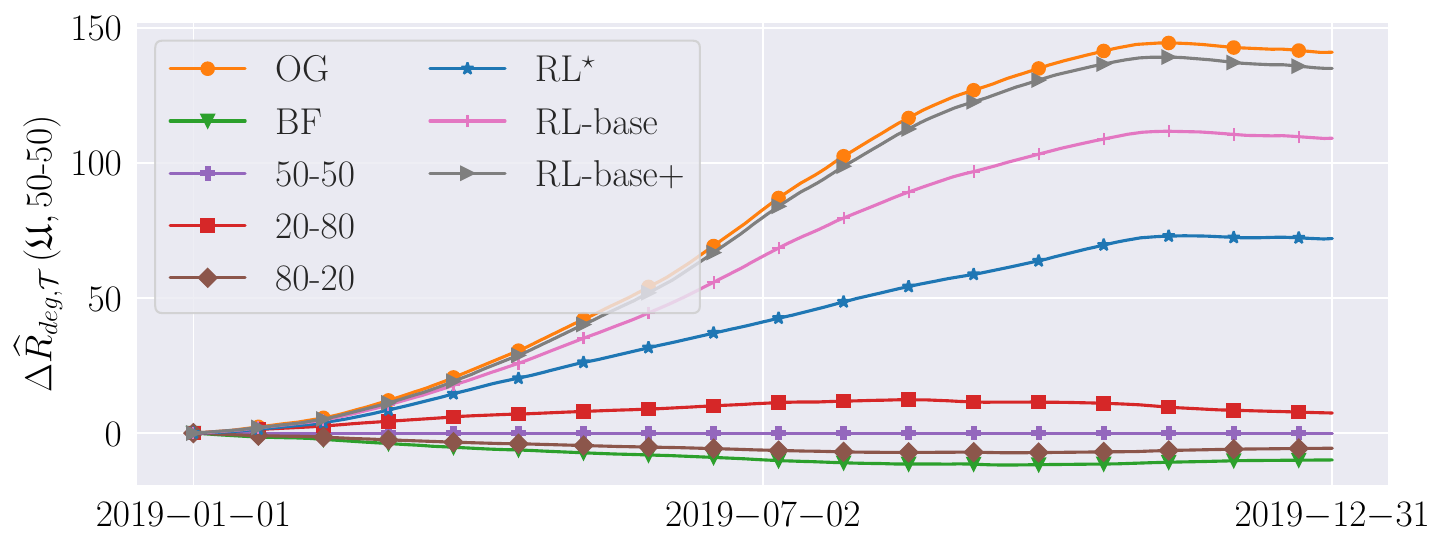}
            \label{fig:r_deg}}
    \end{minipage}

    \caption{Gaps for the trading (a) and degradation (b) components of the reward w.r.t.~the \emph{50-50} policy.}
    \vspace{-0.3cm}
\end{figure}

\begin{figure}[th]
    \vspace{-0.3cm}
    \centering
    \includegraphics[width=0.81\linewidth]{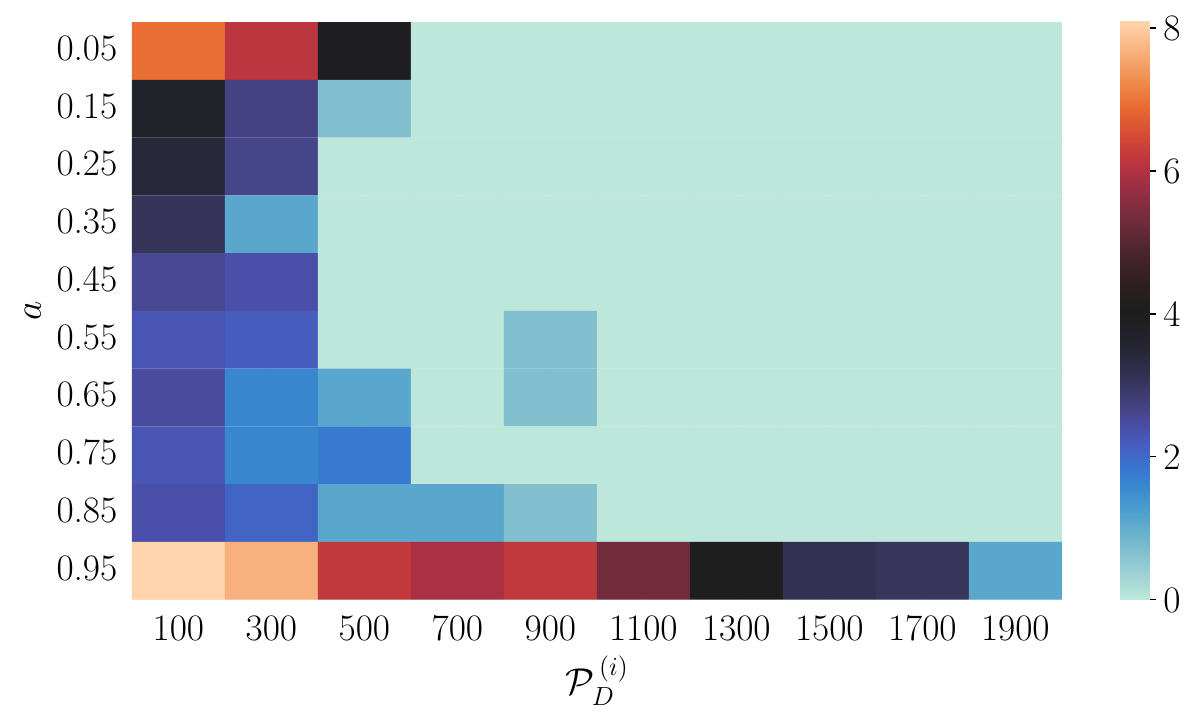}
    \vspace{-0.3cm}
    \caption{Action vs. demand during a test profile run.}
    \label{fig:heatmap}
\end{figure}

By inspecting Figures~\ref{fig:r_trad} and~\ref{fig:r_deg}, we can infer that similar rewards can be achieved with completely different strategies. Let us compare the OG strategy, namely the best among the rule-based ones, with ours. The degradation gap $\Delta \widehat{R}_{deg,t}(OG,\emph{50-50})$ for OG w.r.t.~\emph{50-50} is the largest one, meaning that it is the method using the BESS the least. However, the corresponding trading gap $\Delta \widehat{R}_{trad,t}(OG,\emph{50-50})$ is negatively compensating the previous advantage. Conversely, the $RL^*$ approach balances the two gaps, highlighting a substantial usage of the BESS. 
This can also be observed in Figure~\ref{fig:heatmap}, which presents a heatmap showing how $RL^\star$ plays different actions in response to varying demand levels throughout the test year. Each tile represents the logarithm of the number of times an action is chosen.\footnote{For visualization reasons, we report the logarithm of the number of occurrences of the actions.} This figure highlights that as the demand rises, the action progressively increases the amount of power the algorithm is addressing to the BESS.
Using such a strategy, $RL^\star$ obtains a larger overall reward than OG. These results suggest that trivial rule-based approaches might not be flexible enough for this setting since the best performances are achieved using a strategy that can adapt to external environmental changes.

Now, let us focus on the comparison of the three RL-based approaches. Even if their empirical rewards are similar, by inspecting their gaps in Figures~\ref{fig:r_trad} and~\ref{fig:r_deg}, we infer that \emph{RL-base} and \emph{RL-base+} tend to trade more with the market than $RL^\star$ (smaller trade gap), and they make less intensive usage of the BESS (larger degradation gap). This phenomenon is more evident for \emph{RL-base+}, whose policy is almost equal to the OG. This might be because it employs approximate information about the degradation, overestimating the associated cost. This highlights how the joint use of information about the energy market and the ambient temperature, is crucial to allow $RL^\star$ to learn effective strategies for the MG.



\begin{figure}[th]
    \vspace{-0.3cm}
    \centering
    \includegraphics[width=\linewidth]{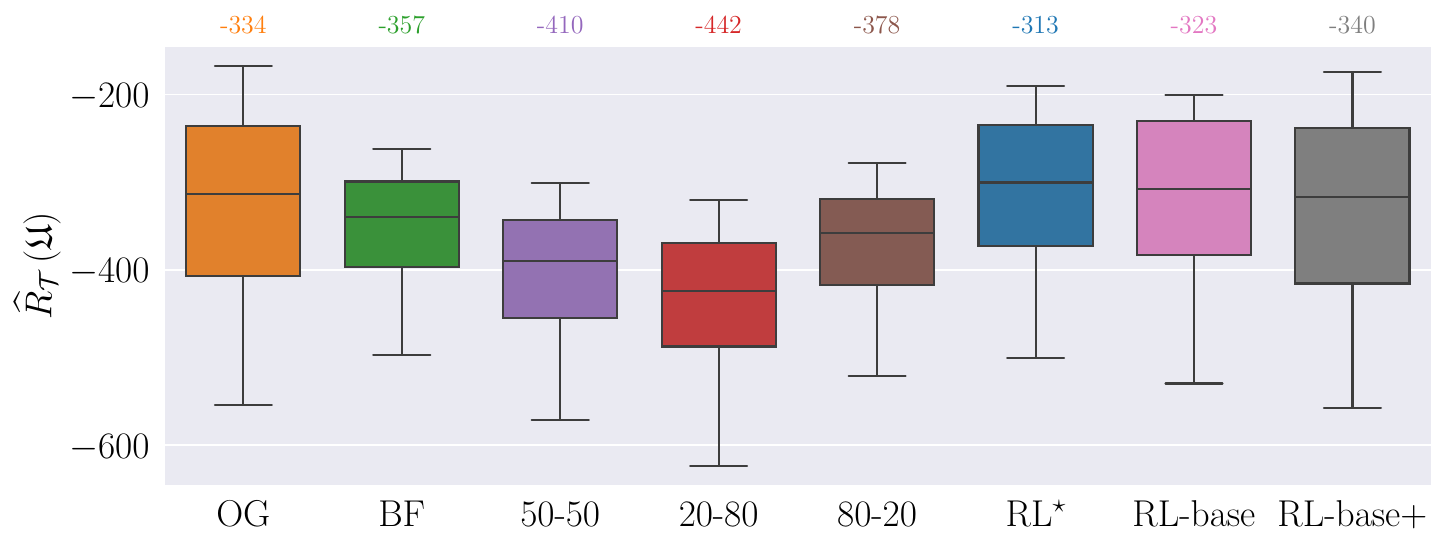}
    \vspace{-0.5cm}
    \caption{Boxplot representing the empirical return across the $28$ validation profiles. Values above the plot represent the return of each method averaged over test profiles.}
    \label{fig:boxplot}
    \vspace{-0.3cm}
\end{figure}

In Figure~\ref{fig:boxplot}, we provide boxplots representing the empirical distribution of the reward $\widehat{R}_{\square,t}(\mathfrak{U})$ at the end of time horizon $\mathcal{T}$ of the considered strategies over the different demand profiles.
The figure shows that returns significantly fluctuate across testing profiles. 
This is because the yearly demand for different profiles spans from $1.50$ to $5.05$ MWh over the test set.
However, we can see that the median of the reward for $RL^*$ is close to the $75\%$ percentiles of most rule-based algorithms and has lower variability w.r.t.~the OG one. Even when compared with the other RL-based approaches, $RL^*$ provides the largest median value for the reward and the smallest spread (having a smaller box and shorter whiskers). This corroborates the idea that the proposed approach can provide more consistent results over the considered profiles.
Finally, a paired t-test to check if the difference between the rewards of $RL^*$ and those of the other methods is significantly greater than zero provided us with a p-value (overall) of $0.0052$, proving that there is strong statistical significance that our method is consistently performing better than the others.

\subsection{Adaptability to market and battery cost shifts}

We evaluated the robustness of our approach across various scenarios with modified energy prices and battery costs. In particular, we performed experiments in which we considered a multiplicative term $\alpha > 0$ for both the selling and buying prices, i.e., we used $\alpha \mathcal{C}_{sell}$ and $\alpha \mathcal{C}_{buy}$ in place of $\mathcal{C}_{sell}$ and $\mathcal{C}_{buy}$, respectively (during the training and testing of our algorithms). This modification mimics the increase ($\alpha > 1$) or decrease ($\alpha < 1$) of the energy price caused by geopolitical events or natural catastrophes, e.g., the  Covid-19~\cite{krarti2021review}. 

\begin{figure}[th]
    \vspace{-0.3cm}
    \centering
    \includegraphics[width=\linewidth]{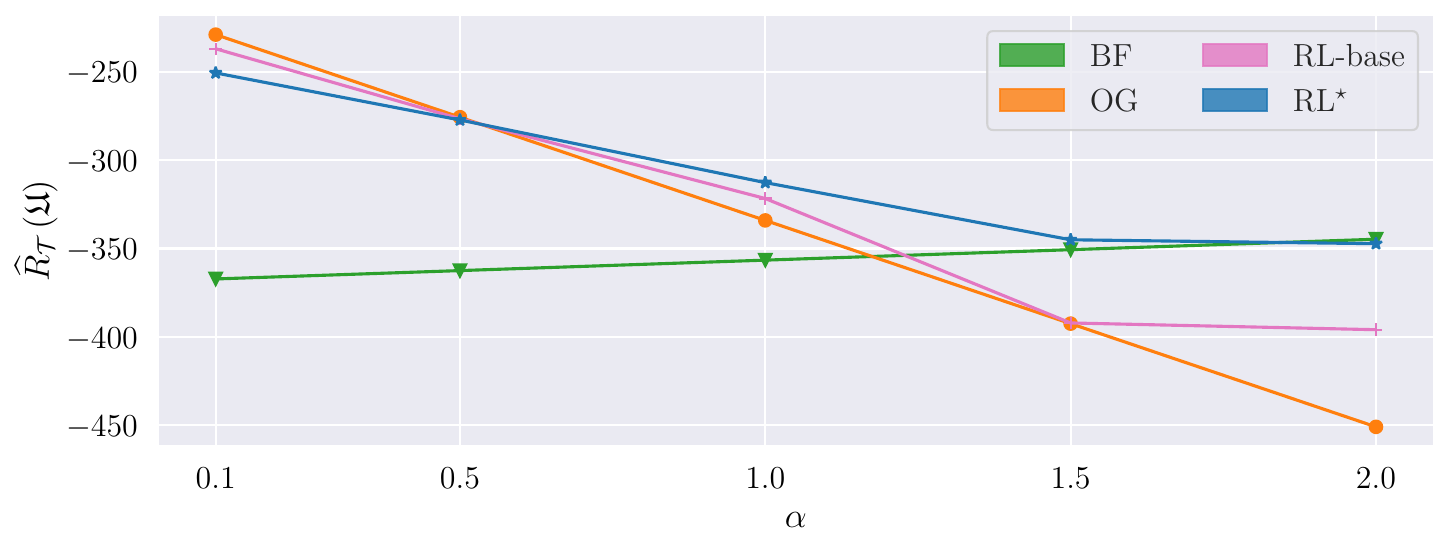}
    \vspace{-0.5cm}
    \caption{Reward at the end time horizon for different $\alpha$.}
    \label{fig:dichotomy}
    \vspace{-0.5cm}
\end{figure}

\begin{figure}[th]
    \vspace{-0.3cm}
    \centering
    \includegraphics[width=\linewidth]{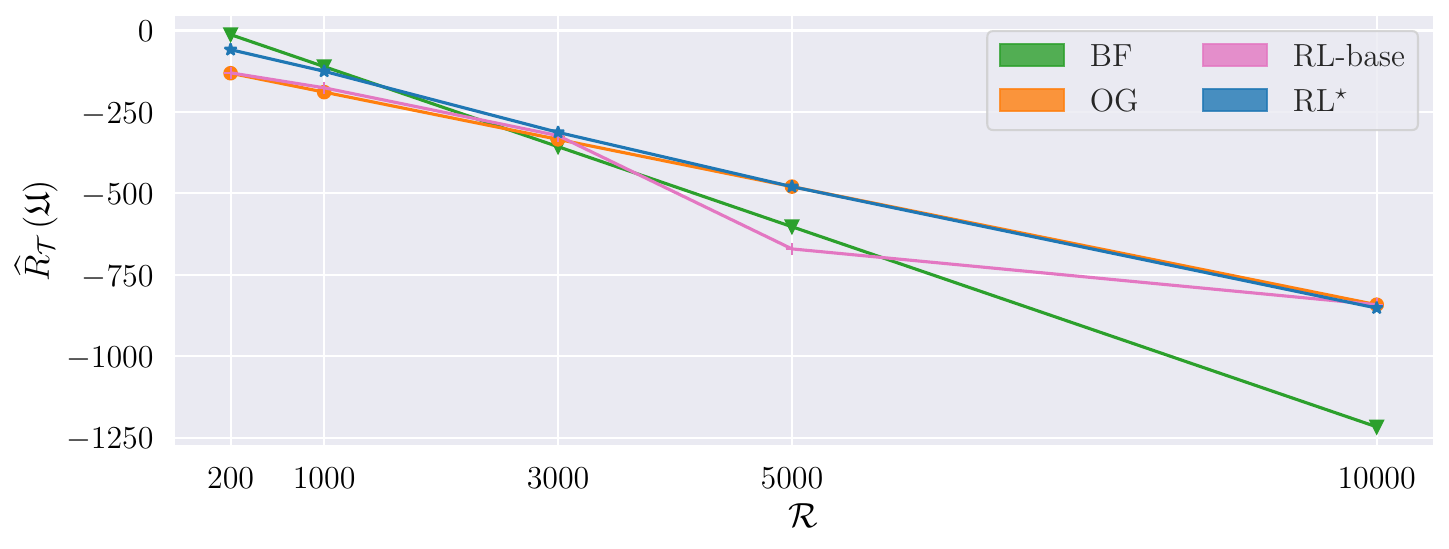}
    \vspace{-0.5cm}
    \caption{Reward at the end time horizon for different $\mathcal{R}$.}
    \label{fig:replacement_cost}
    \vspace{-0.5cm}
\end{figure}

Figure~\ref{fig:dichotomy} show the results of the reward $\widehat{R}_{\mathcal{T}}(\mathfrak{U})$ with different values of the parameter $\alpha \in \{0.1, 0.5, 1, 1.5, 2\}$.
When the energy cost is small, the most profitable strategy is the \emph{OG}, since the financial loss from market transactions diminishes. Conversely, as $\alpha$ increases, the optimal strategy shifts toward \emph{BF}, since purchasing energy becomes prohibitively expensive, making the battery degradation a more cost-effective choice.
Moving to RL strategies, $RL^\star$ outperforms \emph{RL-base} for $\alpha > 0.5$ and is on par with all the other approaches for all the values of $\alpha$.
Figure~\ref{fig:replacement_cost} presents a similar analysis, examining the impact of varying the replacement cost $\mathcal{R} \in \{200, 1000, 3000, 5000, 10000\}$~€. The results confirm the same insights regarding  $RL^*$ behavior as those observed in the experiment on $\alpha$ variation.
These results, regarding both market and replacement cost alterations, corroborate that our approach can adapted to different environmental conditions.

\subsection{Ablation study for clipping parameter}
We study the influence of the clipping term on the learning of the optimal policy.
We compare the reward at the end of the time horizon $\widehat{R}_{\mathcal{T}}(\mathfrak{U})$ for different values of the parameter, specifically $\lambda \in \{ 0, 0.01, 0.05, 0.1, 0.25, 0.5, 1, 2\}$.

\begin{figure}[th!]
    \vspace{-0.3cm}
    \centering
    \includegraphics[width=\linewidth]
    {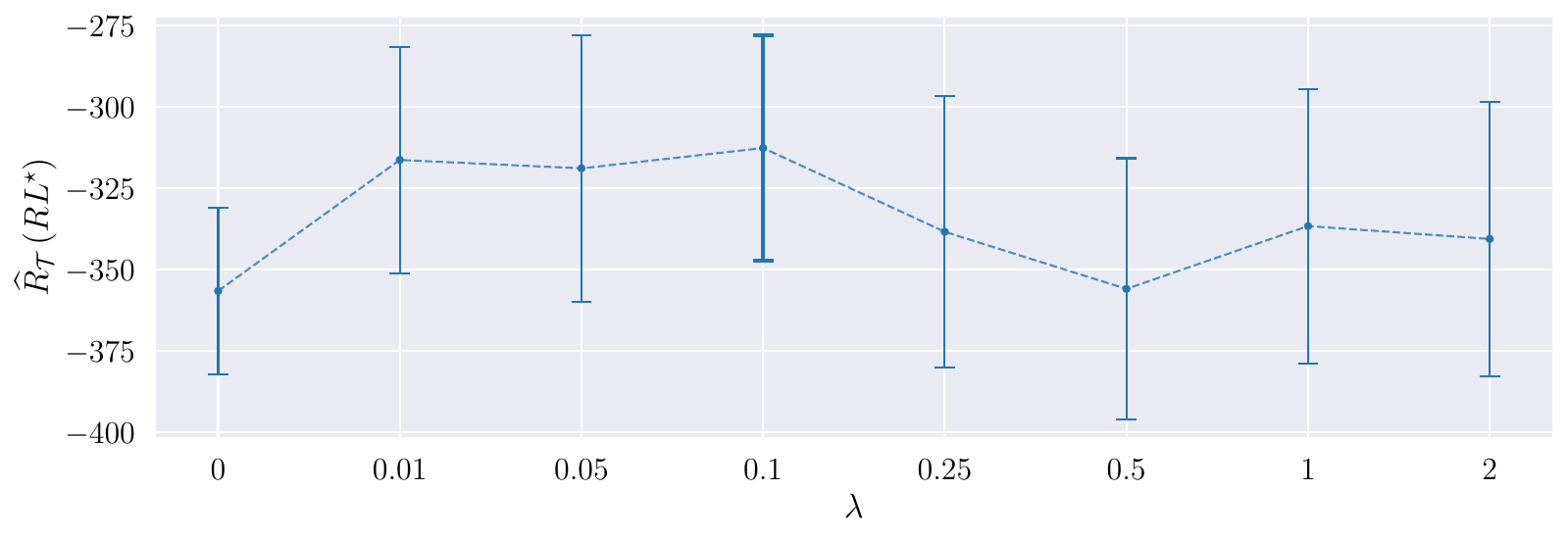}
    \vspace{-0.5cm}
    \caption{Clipping term analysis.}
    \label{fig:clipping}
    \vspace{-0.5cm}
\end{figure}

The optimal value chosen for $RL^\star$ is $\lambda=0.1$. Figure~\ref{fig:clipping} shows that, on average, the configuration with $\lambda = 0.1$ (the one used for the main experiment) provides the best results. However, even using values far from the optimal one, the average reward is still larger than the one of most of the rule-based strategies and is below $6.7 \%$ to the OG one. This shows how, to achieve optimal performance, you must carefully tune the $\lambda$ parameter, but misspecification of such a parameter is not too crucial to obtaining significant results.

\section{Conclusion and Future Works} \label{sec:6}

This paper studies the problem of optimizing MG energy management, a task become crucial nowadays due to the paradigm shift driven by the rapid diffusion of DERs. The availability of historical datasets and the development of DT for BESS allowed the employment of advanced ML approaches to effectively address this challenge. We propose a novel RL-based approach, namely $RL^*$, able to learn from historical data a control policy that maximizes the cumulated reward 
of MG management. The proposed approach was evaluated in a real-world scenario involving Italian MGs. Our experimental validation confirms the effectiveness of the approach, demonstrating its superiority over traditional rule-based methods and prior RL benchmarks.
Despite its promising results, several challenges still remain, the most relevant being the integration of our approach with multi-agent logic~\cite{gao2023distributed, sui2020multi-agent} to enable cooperative energy trading among interconnected MGs.

\vspace{-2pt}

\section*{Acknowledgment}
This work has been funded by the Research Fund for the Italian Electrical System under the Three-Year Research Plan 2022-2024 (DM MITE n. 337, 15.09.2022), in compliance with the Decree of April 16th, 2018. This paper is supported by PNRR-PE-AI FAIR project funded by the NextGeneration EU program.

\bibliographystyle{IEEEtran}
\bibliography{biblio}

\end{document}